\documentclass[lettersize,journal]{IEEEtran}
\usepackage{amsmath,amsfonts}
\usepackage{algorithmic}
\usepackage{algorithm}
\usepackage{array}
\usepackage[caption=false,font=normalsize,labelfont=sf,textfont=sf]{subfig}
\usepackage{textcomp}
\usepackage{stfloats}
\usepackage{url}
\usepackage[dvipsnames]{xcolor}
\usepackage{hyperref}
\usepackage{verbatim}
\usepackage{graphicx}
\usepackage{cite}
\usepackage{multirow}
\usepackage[flushleft]{threeparttable}
\begin{document}

\title{Facial Landmark Detection Evaluation on MOBIO Database}
\author{Na Zhang
\thanks{Na Zhang is with Lane Department of Computer Science and Electrical Engineering at West Virginia University, Morgantown, WV 26506-6109. }
}



\maketitle

\begin{abstract}
MOBIO is a bi-modal database that was captured almost exclusively on mobile phones. It aims to improve research into deploying biometric techniques to mobile devices. Research has been shown that face and speaker recognition can be performed in a mobile environment. Facial landmark localization aims at finding the coordinates of a set of pre-defined key points for 2D face images. A facial landmark usually has specific semantic meaning, e.g. nose tip or eye centre, which provides rich geometric information for other face analysis tasks such as face recognition, emotion estimation and 3D face reconstruction. Pretty much facial landmark detection methods adopt still face databases, such as 300W, AFW, AFLW, or COFW, for evaluation, but seldomly use mobile data. Our work is first to perform facial landmark detection evaluation on the mobile still data, i.e., face images from MOBIO database. About 20,600 face images have been extracted from this audio-visual database and manually labeled with 22 landmarks as the groundtruth. Several state-of-the-art facial landmark detection methods are adopted to evaluate their performance on these data. The result shows that the data from MOBIO database is pretty challenging. This database can be a new challenging one for facial landmark detection evaluation.
\end{abstract}


\section{Introduction}
\label{intro}
\par The mobile biometrics database, MOBIO \cite{mccool2012bi}, is an audio-visual database captured almost exclusively using mobile phones. It is taken from 152 persons with 100 males and 52 females, and collected from August 2008 until July 2010 in six different sites from five different countries with both native and non-native English speakers. This mobile phone database consists of over 61 hours of audio-visual data with 12 distinct sessions usually separated by several weeks. One special point is that the acquisition device is given to the user, rather than being in a fixed position, which makes this database unique and now being used in an interactive and uncontrolled manner. The MOBIO database provides a challenging test-bed for face verification, speaker verification, and bi-modal verification. 

\par Facial landmark detection, also known as face alignment or facial landmark localization, is a mature field of research. In recent years, facial landmark detection has become a vary active area, due to its importance to a variety of image and video-based face analysis systems, such as face recognition \cite{liu2017sphereface, masi2016pose, taigman2014deepface, yang2017neural}, facial expression analysis \cite{fabian2016emotionet, li2017reliable, walecki2016copula, zeng2008survey}, human-computer interaction, video games and 3D face reconstruction \cite{dou2017end, kittler20163d, huber2016real, hu2017efficient, roth2016adaptive, koppen2018gaussian}. Hence, accurate face landmarking and facial feature detection is an important intermediary step for many subsequent face processing operations that range from biometric recognition to the understanding of mental states, which have an impact on subsequent tasks focused on the face, such as coding, face recognition, expression and/or gesture understanding, gaze detection, animation, face tracking, etc.

\par Since face alignment is essential to many face applications, the requirement for the efficiency of facial landmark detection becomes higher and higher, especially when more face images and videos captured in the wild appear. Hence, the large visual variations of faces, such as occlusions, large pose variations and extreme lightings, impose great challenges for face alignment in real world applications. For facial landmark detection evaluation, still face images, like 300W, AFW, AFLW, are universally used. However, mobile face data, e.g., the MOBIO database, is seldomly adopted for facial landmark evaluation so far. In this work, we try to perform facial landmark detection on the mobile still face data using up-to-date methods, and check their performance on this type of faces.

\par A total of 20,600 still face images are extracted from MOBIO database and labelled manually with 22 facial feature points as groundtruth. Seven state-of-the-art facial landmark localization methods are chosen to perform facial landmark detection on these face images. And several measure metrics, e.g., Normalized Mean Error (NME), Cumulative Error Distribution (CED), Area-Under-the-Curve (AUC) and failure rate, are calculated or drawn for evaluation. The experimental result shows that these mobile still face images are pretty challenging for existing facial landmark detection technology and could be a new database for facial landmark localization. This evaluation could establish baseline performance for the MOBIO mobile face images. 

\par The contributions of our work includes:
\begin{itemize}
\renewcommand\labelitemi{\textbullet}
\item generate a still mobile face database with a total of 20,600 images based on video-visual database MOBIO with 22 manually labelled facial landmarks as groundtruth;
\item adopt seven state-of-the-art facial landmark detection methods to evaluate their performance on these 20,600 face images;
\item the result shows that the mobile faces in MOBIO is pretty challenging which can be used as a new database for facial landmark detection evaluation in mobile condition.
\end{itemize}

\par This paper is organized as follows. In section \ref{landmark}, we briefly describe facial landmark detection technique. In section \ref{database}, the still faces based on MOBIO are generated and labeled with 22 facial landmarks. Our approach procedure is given in section \ref{approach}. And experimental results are provided in section \ref{exp}. In section \ref{con}, some interesting discussion and conclusions are drawn.


\section{Facial Landmark Detection}
\label{landmark}
\par In this section, we talk about some basic information about facial landmark detection, the challenges it faces, several categories of detection methods, types of facial points and measure metrics.  

\begin{figure*}[!t]
\center
\includegraphics[width=1.0\linewidth]{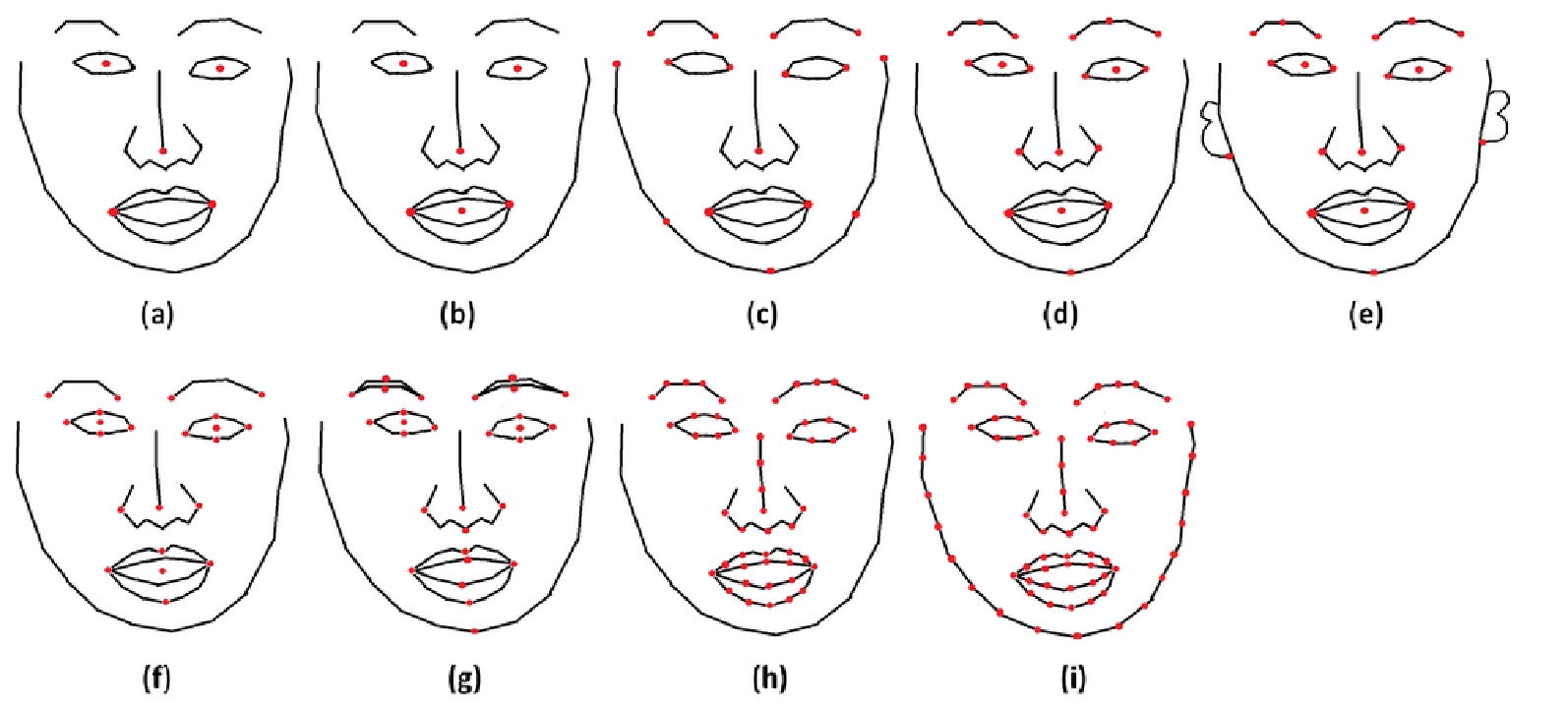}
\caption{Types of facial landmarks with (a) 5, (b) 6, (c) 16, (d) 19, (e) 21, (f) 22, (g) 29, (h) 49, (i) 68 points.} 
\label{fig:typep}
\end{figure*}

\subsection{What is Facial Landmark Detection?}
\par Facial landmark detection, or facial landmark localization, or face alignment, is to automatically localize a set of pre-defined semantic key points including eyes, nose, mouth and other points on the 2D face images. A facial landmark usually has specific semantic meaning, e.g. nose tip or eye center, which provides rich geometric information for other face analysis tasks such as face recognition \cite{taigman2014deepface, masi2016pose, liu2017sphereface, yang2017neural}, emotion estimation \cite{fabian2016emotionet, li2017reliable, walecki2016copula, zeng2008survey} and 3D face reconstruction \cite{dou2017end, kittler20163d, huber2016real, hu2017efficient, roth2016adaptive, koppen2018gaussian}. It is a fundamental problem in computer vision study and an essential initial step for a number of research areas, and plays a key role in many face processing applications, including head pose estimation \cite{demirkus2015hierarchical, zhu2012face}, facial expression analysis and emotion recognition \cite{ding2013facial,  martinez2016advances, sariyanidi2014automatic}, face attribute analysis \cite{liu2015deep, cristinacce2006feature}, face alignment in 2D \cite{cao2014face, xiong2013supervised} and 3D (e.g., frontalization \cite{hassner2015effective}, face 3D modeling, video games, multimodal sentiment analysis \cite{zadeh2016mosi}, person identification, and, of course, face recognition (see, e.g., Sun et al. \cite{sun2014deep} and many others). 

\subsection{ What is the Challenge?}
\par Due to its relevance to many facial analysis tasks, facial landmark detection has attracted increasing interests in the past couple of years. It is a well-researched problem with large amounts of annotated data, and impressive progress has been made too. Current methods could provide reliable results for near-frontal face images \cite{cao2014face, xiong2013supervised, ren2014face, zhang2014coarse, zhu2015face, zhang2015learning}. Thanks to the successive developments in this area of research during the past decades, facial landmark localization can be performed very accurately in constrained scenarios, even using traditional approaches such as Active Shape Model (ASM) \cite{cootes1995active}, Active Appearance Model (AAM) \cite{cootes2001active} and Constrained Local Model (CLM) \cite{cristinacce2006feature}. As the rapid development of deep learning technology, facial landmark detection gains a pretty good performance in unconstrained environment. 

\par Though great strides have been made in this field, facial landmark detection is particularly daunting considering the real-world, unconstrained imaging conditions. In an uncontrolled setting, face is likely to have large out-of-plane tilting, occlusion, illumination and expression variations. Robust facial landmark detection remains a formidable challenge in the presence of partial occlusion and large head pose variations. Images often portray faces in myriads of poses, expressions, occlusions and more, any of which can affect landmark appearances, locations or even presence. Therefore, it is still a challenging problem for localizing landmarks in face images with partial occlusions or large appearance variations due to illumination conditions, poses, and expression changes. 

\subsection{Categories of Methods}
\par In general, existing facial landmark detection methods can be divided into two categories: (1) traditional approaches, e.g., ASM \cite{cootes1995active} and AAM \cite{cootes2001active} based methods, which fit a generative model by global facial appearance; (2) cascade regression based methods, which try to estimate the facial landmark positions by a sequence of regression models. In recent year, deep learning based cascade regression models have performed robust facial landmark localization using deep neural networks.

\par \textbf{ASM and AAM based methods.} This kind of methods is traditional approaches, which usually perform accurately in constrained scenarios. They rely on a generative PCA-based shape model. However, these methods require expensive iterative steps and rely on good initialization. The mean shape is often used as the initialization, which may be far from the target position and hence inaccurate.


\par \textbf{Cascade regression based methods.} In cascade regression framework, a set of weak regressors are cascaded to form a strong regressor \cite{wu2017simultaneous, feng2017face, wu2016constrained, feng2014random, cao2014face, xiong2013supervised}. It tries to obtain the coarse location first, and the following steps are to refine the initial estimate, yielding more accurate results. Cascade regression directly positions facial landmarks on their optimal locations based on image features. The shape update is achieved in a discriminative way by constructing a mapping function from robust shape related local features to shape updates. However, these methods need to train individual systems for each group of the landmarks, the computational burden grows proportional to the group numbers and cascade levels. For example, the cascaded Convolutional Neural Network (CNN) method \cite{sun2013deep} needs to train 23 individual CNN networks. However, the capability of cascaded regression is nearly saturated due to its shallow structure. After cascading more than four or five weak regressors, the performance of cascaded regression is hard to improve further \cite{feng2015cascaded}.

\par More recently, as deep neural networks have been put forward as a more powerful alternative in a wide range of computer vision and pattern recognition tasks, facial landmark localization gains large development too. Different network types have been explored, such as Convolutional Neural Network (CNN), Auto-Encoder Network and Recurrent Neural Network, to perform robust facial landmark localization. In our work, most of methods adopted belong to deep learning based models, such as Wing loss based method WingLoss \cite{feng2018wing}.

\subsection{Types of Facial Landmarks}
\par Existing facial landmark detection methods can figure out different numbers of facial feature points, e.g., 5, 6, 16, 19, 21, 22, 29, 49, 68, etc. Figure \ref{fig:typep} gives several typical facial landmarks. Figure \ref{fig:typep}(a) consists five feature points(i.e., left eye center, right eye center, nose tip, left mouth corner, and right mouth corner). Figure \ref{fig:typep}(b) is a face with six points with one more landmark, mouth center, than Figure \ref{fig:typep}. Besides feature points of eye area, nose, and mouth, Figure \ref{fig:typep}(c) considers five face contour landmarks. Figure \ref{fig:typep}(d) and (e) share similar landmarks, the only difference is Figure \ref{fig:typep}(e) have two extra points on two ears. Figure \ref{fig:typep}(f)~(i) provide more points to describe geometric information of face.




\subsection{Landmark Performance and Evaluation Metrics}
\par There are two different metrics to evaluate landmark detection performance, task-oriented performance and ground-truth based localization performance. For task-oriented performance, one can measure the impact of the landmark detection accuracy on the performance scores of a task. For ground-truth based localization performance, a straightforward way is to use manually annotated ground-truths. 

\par In practice, ground-truth based localization performance is commonly used in facial landmark detection for thorough analysis. If the ground-truth positions are available, the localization performance can be expressed in terms of the Normalized Mean Error (NME), Cumulative Error Distribution (CED) curve, Area-Under-the-Curve (AUC) and failure rate.

\begin{figure}[!t]
\center
\includegraphics[width=1.0\linewidth]{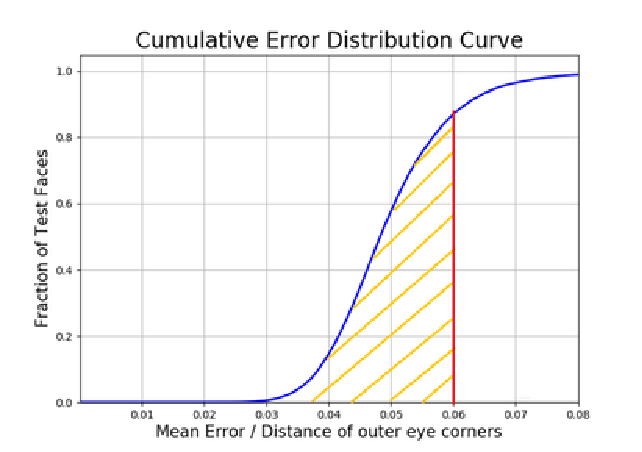}
\caption{An illustration of CED and AUC.} 
\label{fig:cedex}
\end{figure}

\par Normalized Mean Error (NME) is a primary metric in facial landmark detection evaluation. It is calculated first by the distances between the estimated landmarks and the groundtruths, and then normalized with respect to the inter-ocular distance, i.e. Euclidean distance between two eye centres, the distance of outer or inner corners of the eyes. The mean error has three types: landmark-wise, sample-wise or overall. Landmark-wise face alignment error is first normalized in the following way to make it scale invariant: $e_x = \frac{\| \hat{x}-x^{GT}\|}{D_{IOD}}$ where $\| \hat{x}-x^{GT}\|$ is the Euclidean distance between the estimated location $\hat{x}$ and the true location $x^{GT}$. $D_{IOD}$ is the inter-ocular distance (IOD). Normalizing landmark localization errors by dividing with IOD makes the performance measure independent of the actual face size or the camera zoom factor. Sample-wise mean error could be calculated as $ \frac{1}{n} \sum^n_{i=1}{\frac{\| \hat{x_i} - x_i^{GT} \|}{D_{IOD}}}$, where n is the number of facial landmarks involved in the evaluation. The error is normalized by the distance of outer or inner corners of the eyes. NME can be averaged over all the landmarks to produce a global precision figure, which is a overall mean error. In recent years, with the rapid progress of face alignment, most of the recent approaches report a error level of e at around 0.05 or smaller, which is close to human performance.

\par Using NME is very straightforward and intuitive given its single value form. However, this measure is heavily impacted by the presence of some big failures such as outliers, in particular when the average error level is very low. In other words, the mean error measure is very fragile even if there are just a few images with big errors. Thus though the mean error is widely used for face alignment evaluation \cite{jourabloo2015pose, lee2015face, ren2014face, sun2013deep, zhang2014facial}, it does not provide a big picture on which cases the errors occur, e.g., minor big alignment error, many inaccuracies.


\par Since using overall mean error as an evaluation criterion is too sensitive to big erroneous samples, Cumulative Error Distribution (CED) curve and Area-Under-the-Curve (AUC) are adopted as two better metrics (see Figure \ref{fig:cedex}). CED curve is the cumulative distribution function of the normalized error as shown by the blue line in Figure \ref{fig:cedex}. The x-axis is error value, and y-axis is the fraction of test faces. In terms of outliers handling, CED is a better way. However, it is not intuitive given its curve representation. It is also hard to use it in sensitivity analysis. Therefore, AUC is adopted. It is calculated from the CED curve. The AUC stands for the value of area under the curve of CED.  It is defined as: $AUC_{\alpha}=\int^{\alpha}_0 f(e)de$ where e is the normalized error, f(e) is the cumulative error distribution function and $\alpha$ is the upper bound that is used to calculate the definite integration. In Figure \ref{fig:cedex}, $\alpha$ is 0.06 (red line). Given the definition of the CED function, the value is $AUC_{\alpha}$ lies in the range of [0, $\alpha$], the area with yellow titled lines. The value of $AUC_{\alpha}$ will not be influenced by points with error bigger than $\alpha$. 

\par Landmark detection statistics can be characterized by the exceedance probability of the localization error. A general agreement in the literature is that e < 0.1 is an acceptable error criterion so that a landmark is considered detected whenever it is found within proximity of one tenth of the inter-ocular distance from its true position. In our experiment, $\alpha$ is set to 0.08 and 0.1.

\par Failure rate is calculated with the threshold for the normalized mean error. It computes the fraction of test faces that the error value of which is larger than the threshold. In our experiment, the thresholds are set to 0.1 and 0.08.

\subsection{Common Databases Used}
\par Annotated databases are extremely important in computer vision. Therefore, a number of databases containing faces with different facial expressions, poses, illumination and occlusion variations have been collected in the past. Most evaluation experiments are conducted on commonly used benchmark datasets, such as the 300 Faces in the Wild (300W) \cite{sagonas2013300}, Annotated Facial Landmarks in the Wild (AFLW) \cite{koestinger2011annotated}, the Annotated Faces in-the-wild (AFW) \cite{zhu2012face}, the Labeled Face Parts in-the-wild (LFPW) \cite{belhumeur2013localizing}, HELEN \cite{le2012interactive}, the Caltech Occluded Faces in the Wild (COFW) \cite{burgos2013robust}. The aforementioned databases, cover large variations including: different subjects, poses, illumination, occlusion, etc. 

\par
\begin{figure*}[!t]
\center
\includegraphics[width=1.0\linewidth]{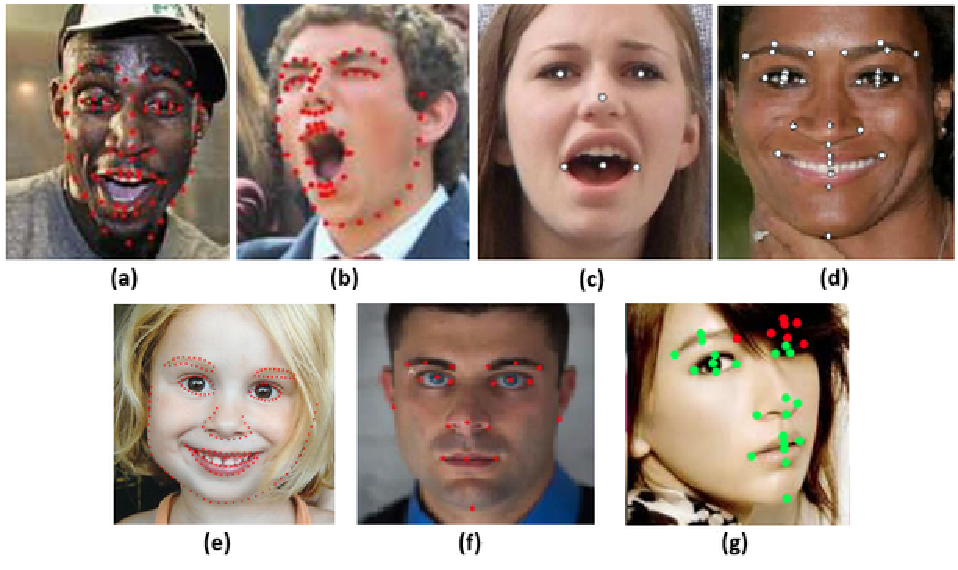}
\caption{Annotated images from (a) 300W (Indoor), (b) 300W (Outdoor), (c) AFW, (d) LFPW, (e) HELEN, (f) AFLW and (g) COFW.} 
\label{fig:annotated}
\end{figure*}

 The 300 Faces in the Wild (300W) \cite{sagonas2013300} dataset is a commonly used benchmark for facial landmark localization problem. It contains near-frontal face images in the wild and provides 68 semi-automatically annotated points for each face. It is created from existing datasets, including LFPW \cite{belhumeur2013localizing}, AFW \cite{zhu2012face}, HELEN \cite{le2012interactive}, XM2VTS \cite{messer1999xm2vtsdb} and IBUG \cite{sagonas2013300}. The 300W training set contains 3,148 training images from AFW, LFPW and HELEN. The common subset of 300W contains 554 test images from LFPW and HELEN. The challenging subset of 300W contains 135 test images from IBUG. The fullset of 300W is the union of the common and challenging subset. The 300W test set contains 600 test images which are provided officially by the 300W competition \cite{sagonas2013300} and said to have a similar distribution to the IBUG dataset. IBUG subset is extremely challenging due to the large variations in face pose, expression and illumination. As the rapid progresses of the study in facial landmark localization in recent years, several methods have reported close-to-human performance on this dataset of which the images that are acquired from unconstrained environments. Figure \ref{fig:annotated} shows annotated (a) indoor and (b) outdoor images of 300W.

\par The Annotated Faces in-the-wild (AFW) \cite{zhu2012face} database is a popular benchmark for facial landmark detection, containing 205 images with 468 faces. A detection bounding box as well as up to 6 visible landmarks are provided for each face. An example of an image taken from the AFW database along with the corresponding annotated landmarks is depicted in Figure \ref{fig:annotated} (c).

\par The Annotated Facial Landmarks in-the-wild (AFLW) \cite{koestinger2011annotated} is a very challenging dataset that has been widely used for benchmarking facial landmark localization algorithms. It contains 25,000 images of 24,686 subjects downloaded from Flickr. The images contain a wide range of natural face poses in yaw (from -90 to 90) and occlusions. Facial landmark annotations are available for the whole database. Each annotation consists of 21 landmark points. The AFLW-Full protocol contains 20,000 training and 4,386 test images, and each image has 19 manually annotated facial landmarks. Figure \ref{fig:annotated} (f) depicts an annotated image from AFLW. 



\par The COFW \cite{burgos2013robust} dataset contains in-the-wild face images with heavy occlusions, including 1,345 face images for training and 507 face images for testing. For each face, 29 landmarks and the corresponding occlusion states are annotated in the COFW dataset. An example of an image taken from the COFW database along with the corresponding annotated landmarks is depicted in Figure \ref{fig:annotated} (g).

\par The Labeled Face Parts in-the-wild (LFPW) \cite{belhumeur2013localizing} database contains 1,432 images downloaded from google.com, fickr.com, and yahoo.com. The images
contain large variations including pose, expression, illumination and occlusion. The provided ground truth consists of 29 landmark points. An example of an image taken from the LFPW database along with the corresponding annotated landmarks is depicted in Figure \ref{fig:annotated} (d).

\par The HELEN \cite{le2012interactive} database consists of 2,330 annotated images collected from the Flickr. The images are of high resolution containing faces of size sometimes greater than 500*500 pixels. The provided annotations are very detailed and contain 194 landmark points. Figure \ref{fig:annotated} (e) depicts an annotated image from HELEN.


\section{Face Images Based on MOBIO Database}
\label{database}
\par The MOBIO is an audio-video database of human faces and voice captured almost exclusively on mobile phones. This database is originally used to evaluate the performance of face and speaker recognition in the context of a mobile environment \cite{mccool2012bi, marcel2010results}. 

\par So far, facial landmark detection technique has been seldomly evaluated in the context of a mobile environment. So it is meaningful to perform facial landmark detection on the mobile faces. The mobile environment was chosen as it provides a realistic and challenging test-bed for face points detection techniques to operate. For instance, the environment is quite complex and there is limited control over the illumination conditions and the pose of the subject for the video. 

\par In our work, we extract still face frames from the MOBIO videos and generate a face images database. This section briefly describes the MOBIO database first, and then introduces the face images extracted from the MOBIO video data, finally talks about how to generate the groundtruth of the faces with 22 feature points. 

\subsection{MOBIO Database}
\par MOBIO \cite{mccool2012bi} database is an unique diverse bi-modal database (audio + video) that was captured almost exclusively on mobile phones. It consists of over 61 hours of audio-visual data with 12 distinct sessions usually separated by several weeks. There are a total of 192 unique audio-video samples for each of the 152 participants. Female-Male ratio is 1:2. This data was captured at 6 different sites over one and a half years with people speaking English. Capturing the data on mobile phones makes this database unique because the acquisition device is given to the user, rather than being in a fixed position. This means that the microphone and video camera are no longer fixed and are now being used in an interactive and uncontrolled manner. This database was captured almost exclusively using mobile phones and aims to improve research into deploying biometrics techniques to mobile devices. 


\par The database was acquired primarily on mobile phones. 12 sessions were captured for each participant. 6 sessions for Phase I and 6 sessions for Phase II. In Phase I, the participants are asked to answer a set of questions which are classified as set responses, read speech from a paper, and free speech. Each session consists of 21 questions: 5 pre-defined set response questions, 1 read speech question and 15 free speech questions. Phase II consists of 11 questions with the question types ranging from short response questions, set speech, and free speech. 

\par All videos are recorded using two mobile devices: one mobile phone (NOKIA N93i) and one laptop computer (standard 2008 MacBook). The laptop was only used to capture part of the first session. The first session consists of data captured on both the laptop and the mobile phone.

\par The publicly-available mobile phone database MOBIO (Source download link: https://www.idiap.ch/dataset/mobio) presents several challenges, including: (1) high variability of pose and illumination conditions, even during recordings, (2) high variability in the quality of speech, and (3) variability in the acquisition environments in terms of acoustics as well as illumination and background.

\subsection{Extracted Face Images and Facial Landmark Groundtruth}
\par
\begin{figure*}[!t]
\center
\includegraphics[width=1.0\linewidth]{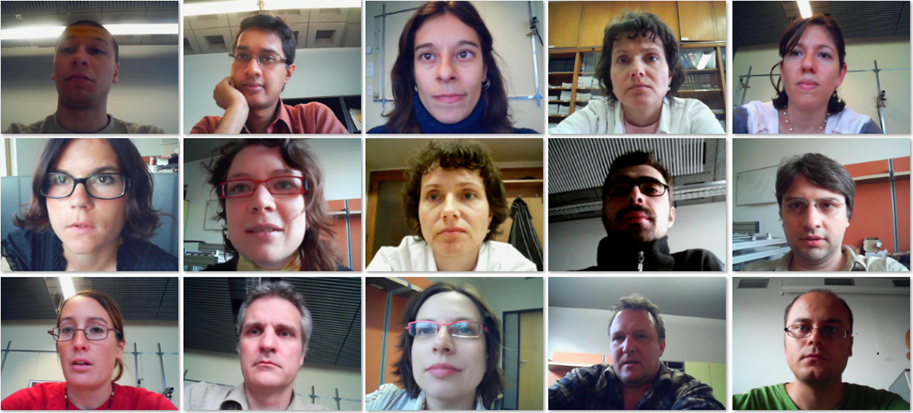}
\caption{Face samples of MOBIO database. }
\label{fig:face}
\end{figure*}

 Based on the video data in MOBIO, a few face frames are extracted for each subject. A total of 20,600 still face images with size of 640*480 are generated finally. The average number of images for each subject is about 136. Figure \ref{fig:face} gives several face samples. Since all images are captured in unconstrained conditions, it contains big variations in head pose, illumination, occlusion (e.g., hair, glass), which makes it a challenging database. 


\par Much work have been done to generate the groundtruth of these face images via manually labeling 22 facial feature points. Figure \ref{fig:landmark} shows the 22 facial landmarks of the face, including 4 points describing brow (left brow left corner, left brow right corner, right brow left corner, right brow right corner), 10 points describing eyes (left eye left corner, left eye top center, left eye right corner, left eye bottom center, left eye center, right eye left corner, right eye top center, right eye right corner, right eye bottom center, right eye center), three points describing nose (nose tip, nose left, nose right), and five points describing mouth (mouth left corner, mouth upper lip center, mouth right corner, mouth bottom lip center, and mouth center).

\begin{figure}[!t]
\center
\includegraphics[width=1.0\linewidth]{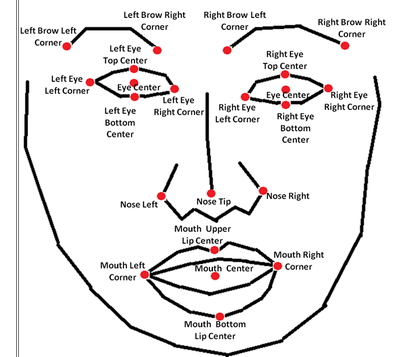}
\caption{22 facial landmarks. }
\label{fig:landmark}
\end{figure}

\par
\begin{figure*}[!t]
\center
\includegraphics[width=1.0\linewidth]{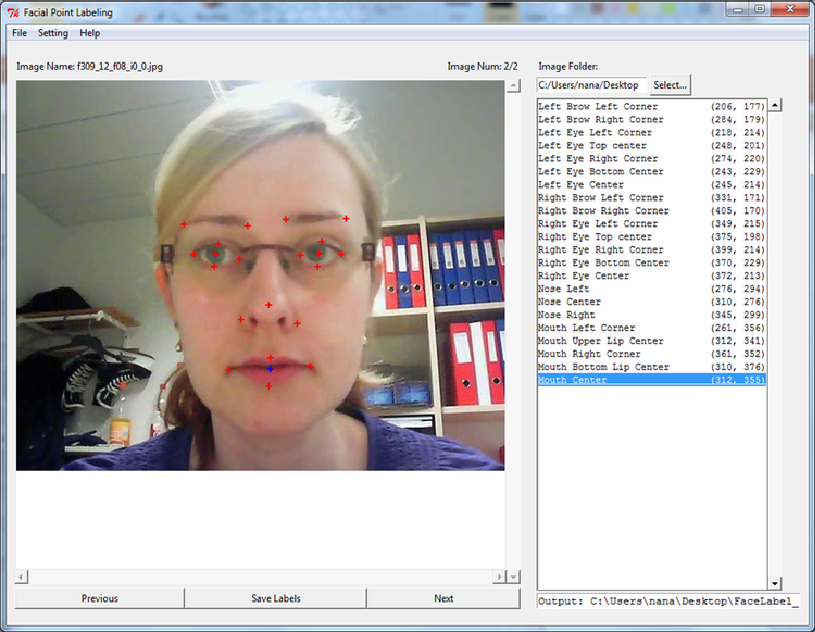}
\caption{Face Label App}
\label{fig:labelapp}
\end{figure*}

 In order to label these faces conveniently and efficiently, a labeling tool named 'Face Label App' as shown in Figure \ref{fig:labelapp} was developed, which can run on Windows system. The users need to load face images first, and then click the 22 facial feature points on each face in a pre-defined order. The app can automatically capture the position (with x, y values) of each facial landmark when mouse moves on it and clicked. All facial landmark position information are saved in .txt files.

\subsection{Preprocess Mobile Still Face Data}
\par Some facial landmark detection methods are able to handle faces with any sizes, like DAC-CSR \cite{feng2017dynamic}, PA-CNN Model \cite{he2016facial}, CE-CLM \cite{zadeh2017convolutional}, etc. However, some detection methods, such as Tweaked CNN \cite{wu2017facial}, WingLoss \cite{feng2018wing}, and ECT \cite{zhang2018combining}, require that their input must be square faces with fixed size (e.g., 256*256). Hence, face detection, cropping and resizing are executed to preprocess the faces for these methods. MTCNN \cite{zhang2016joint} model which is a pretty good face detector is adopted in our work for face detecting. Based on the bounding box of faces, all detected faces are cropped into square shape and then resized to fixed size.





\section{Our Approach}
\label{approach}
\par In this section, we choose several facial landmark detection methods (Tweaked CNN \cite{wu2017facial}, PA-CNN Model  \cite{he2016facial}, WingLoss \cite{feng2018wing}, CE-CLM \cite{zadeh2017convolutional}, ECT \cite{zhang2018combining}, TCDCN \cite{zhang2014facial}, DAC-CSR \cite{feng2017dynamic}) to perform face alignment task and analyze their performance on the mobile still faces and other commonly used face databases like 300W, AFW, AFLW, COFW. Normalized mean error (NME), Cumulative Error Distribution curve (CED), Area Under the error Curve (AUC) and failure rate are adopted as our measure metrics for evaluation.

\subsection{Facial Landmark Detection Methods}
\par We choose seven facial landmark detection methods to detect face feature points on mobile still face images. They are Tweaked CNN \cite{wu2017facial}, WingLoss \cite{feng2018wing}, DAC-CSR \cite{feng2017dynamic}, PA-CNN Model \cite{he2016facial}, OpenPose \cite{zadeh2017convolutional}, ECT \cite{zhang2018combining}, and TCDCN \cite{zhang2014facial}. Most of them are deep learning based methods.

\par Among these deep learning methods, Tweaked CNN \cite{wu2017facial} detects 5 facial points, WingLoss \cite{feng2018wing} and DAC-CSR \cite{feng2017dynamic} detect 19 points, and the others detect 68 points. Some methods of them can do facial landmark detection directly on face images with any sizes. Tweaked CNN \cite{wu2017facial}, WingLoss \cite{feng2018wing} and ECT \cite{zhang2018combining} need face cropping with size of 256*256 before facial landmark detection. MTCNN \cite{zhang2016joint} is adopted in our work to do face detection and cropping due to its efficiency.

\subsection{Methods with 5 Points}
\par Figure \ref{fig:landmark} (a) shows the five facial landmarks (left eye center, right eye center, mouth left corner, mouth right corner, nose tip) that are detected by Tweaked CNN \cite{wu2017facial} models.


\begin{figure*}[!t]
\center
\includegraphics[width=1.0\linewidth]{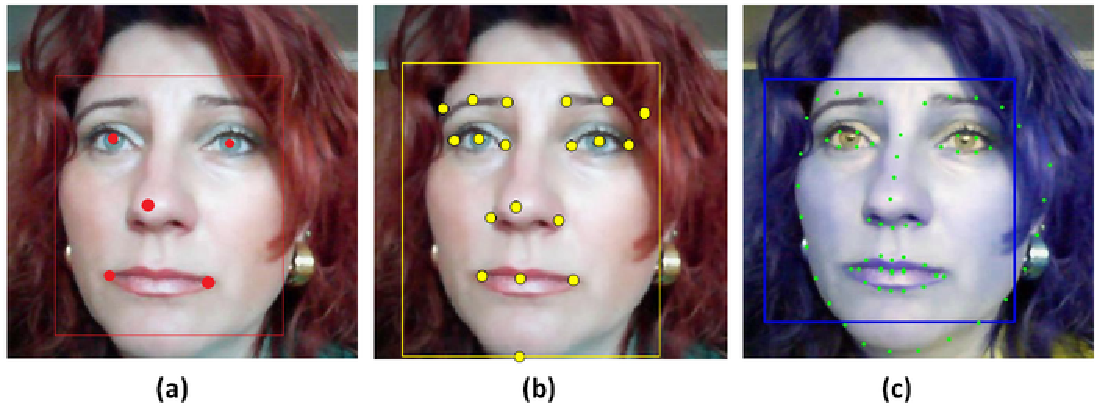}
\caption{5, 19, and 68 Facial landmarks.} 
\label{fig:landmark}
\end{figure*}


\subsubsection{Tweaked CNN Model}
\par Based on the analysis that the features produced at intermediate layers of a convolutional neural network can be trained to regress facial landmark coordinates, face images can be partitioned in an unsupervised manner into subsets containing faces in similar poses (i.e., 3D views) and facial properties (e.g., presence or absence of eye-wear). Therefore, Tweaked CNN (TCNN) \cite{wu2017facial} specializes in regressing the facial landmark coordinates of faces in specific poses and appearances. It is shown to outperform existing landmark detection methods in an extensive battery of tests on the AFW, ALFW, and 300W benchmarks.





\subsection{Methods with 19 Points}
\par Figure \ref{fig:landmark} (b) shows the 19 facial landmarks containing 6 points on brow (left brow left corner, left brow center, left brow right corner, right brow left corner, right brow center, right brow right corner), 6 points on eyes (left eye left corner, left eye center, left eye right corner, right eye left corner, right eye center, right eye right corner), 3 points on nose (nose left, nose tip, nose right), 3 points on mouth (mouth left corner, mouth center, mouth right corner), and 1 point on chin (lower chin center) used by WingLoss \cite{feng2018wing} and DAC-CSR \cite{feng2017dynamic} models.



\subsubsection{WingLoss Model}
\par WingLoss \cite{feng2018wing} method presents a piece-wise loss function, namely Wing loss, for robust facial landmark localization in the wild with Convolutional Neural Networks (CNNs). The loss function pays more attention to small and medium range errors and amplifies the impact of errors from the interval (-w,w) by switching from $L_1$ loss to a modified logarithm function. The experimental results obtained on the AFLW (AFLW-Full protocol) and 300W datasets demonstrate the merits of the Wing loss function, and prove the superiority of the proposed method over the state-of-the-art approaches. 



\subsubsection{DAC-CSR Model}
\par DAC-CSR \cite{feng2017dynamic}, namely Dynamic Attention-Controlled Cascaded Shape Regression architecture, is for robust facial landmark detection on unconstrained faces. It divides facial landmark detection into three cascaded sub-tasks: face bounding box refinement, general cascaded shape regression and attention-controlled cascaded shape regression. The first two stages refine initial face bounding boxes and output intermediate facial landmarks. Then, an online dynamic model selection method is used to choose appropriate domain-specific cascaded shape regressions for further landmark refinement. The key innovation of the DAC-CSR is the fault-tolerant mechanism, using fuzzy set sample weighting, for attention-controlled domain-specific model training. It uses two challenging face datasets, AFLW and COFW, to evaluate the performance of the DAC-CSR architecture.





\subsection{Methods with 68 Points}
\par
\begin{figure}[!t]
\center
\includegraphics[width=1.0\linewidth]{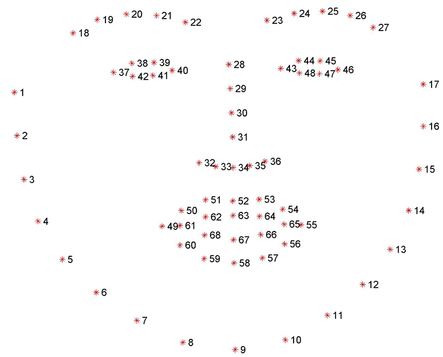}
\caption{68 facial landmarks.} 
\label{fig:68face}
\end{figure}

 Figure \ref{fig:landmark} (c) shows the 68 facial landmarks including 17 contour landmarks and 51 inner landmarks. As shown in Figure \ref{fig:68face}, number 1-17 of points denote the contour landmarks and number 18-68 points denote the inner landmarks.

\subsubsection{PA-CNN Model}
\par PA-CNN Model \cite{he2016facial} is short for Part-Aware Deep Convolutional Neural Network. It is an end-to-end regression framework for facial landmark
localization. It encodes images into feature maps shared by all landmarks. Then, these features are sent into two independent sub-network modules to regress contour landmarks and inner landmarks, respectively. It incorporates the contour landmark sub-network and the inner landmark sub-network into a unified architecture. Contrary to others, this method does not involve multiple individual models or require auxiliary labels. More importantly, the framework treats landmarks on different facial part differently which helps to learn discriminative features. This method can directly detect landmarks on original images. It does not need face detection, cropping, and resizing.  Extensive evaluations are conducted on 300W benchmark dataset.





\subsubsection{CE-CLM Model}
\par Constrained Local Models (CLMs) are a well-established family of methods for facial landmark detection. However, they have recently fallen out of favor to cascaded regression-based approaches. This is in part due to the inability of existing CLM local detectors to model the very complex individual landmark appearance that is affected by expression, illumination, facial hair, makeup, and accessories. 
 
\par CE-CLM \cite{zadeh2017convolutional} introduced a member of CLM family, Convolutional Experts Constrained Local Model (CE-CLM), in which it uses a local detector called Convolutional Experts Network (CEN). CEN brings together the advantages of neural architectures and mixtures of experts in an end-to-end framework. It is able to learn a mixture of experts that capture different appearance prototypes without the need of explicit attribute labeling, and is able to deal with varying appearance of landmarks by internally learning an ensemble of detectors, thus modeling landmark appearance prototypes. This is achieved through a Mixture of Expert Layer, which consists of decision neurons connected with non-negative weights to the final decision layer. Convolutional Experts Constrained Local Model (CE-CLM) algorithm consists of two main parts: response map computation using Convolutional Experts Network and shape parameter update. CE-CLM is able to perform well on facial landmark detection and is especially accurate and robust on challenging profile images.

\subsubsection{ECT Model}
\par The three-step framework named ECT (Estimation-Correction-Tuning) \cite{zhang2018combining} is an effective and robust approach for facial landmark detection by combining data- and model-driven methods. Firstly, a Fully Convolutional Network (FCN) which is a data-driven method is trained to compute response maps of all facial landmark points, which makes full use of holistic information in a facial image for global estimation of facial landmarks. After that, the maximum points in the response maps are fitted with a pre-trained Point Distribution Model (PDM) to generate the initial facial shape. This model-driven method is able to correct the inaccurate locations of outliers by considering the shape prior information. Finally, a weighted version of Regularized Landmark Mean-Shift (RLMS) is employed to fine-tune the facial shape iteratively. 


\par This Estimation-Correction-Tuning process perfectly combines the advantages of the global robustness of data-driven method (FCN), outlier correction capability of model-driven method (PDM) and non-parametric optimization of RLMS. The method is able to produce satisfying detection results on face images with exaggerated expressions, large head poses, and partial occlusions.

\subsubsection{TCDCN Model}
\par Facial landmark detection has long been impeded by the problems of occlusion and pose variation. Instead of treating the detection task as a single and independent problem, TCDCN \cite{zhang2014facial} investigate the possibility of improving detection robustness through multi-task learning. This tasks-constrained deep model can facilitate learning convergence with task-wise early stopping. It optimizes the facial landmark detection together with heterogeneous but subtly correlated tasks, e.g.head pose estimation and facial attribute inference.




\begin{figure*}[!t]
\center
\includegraphics[width=1.0\linewidth]{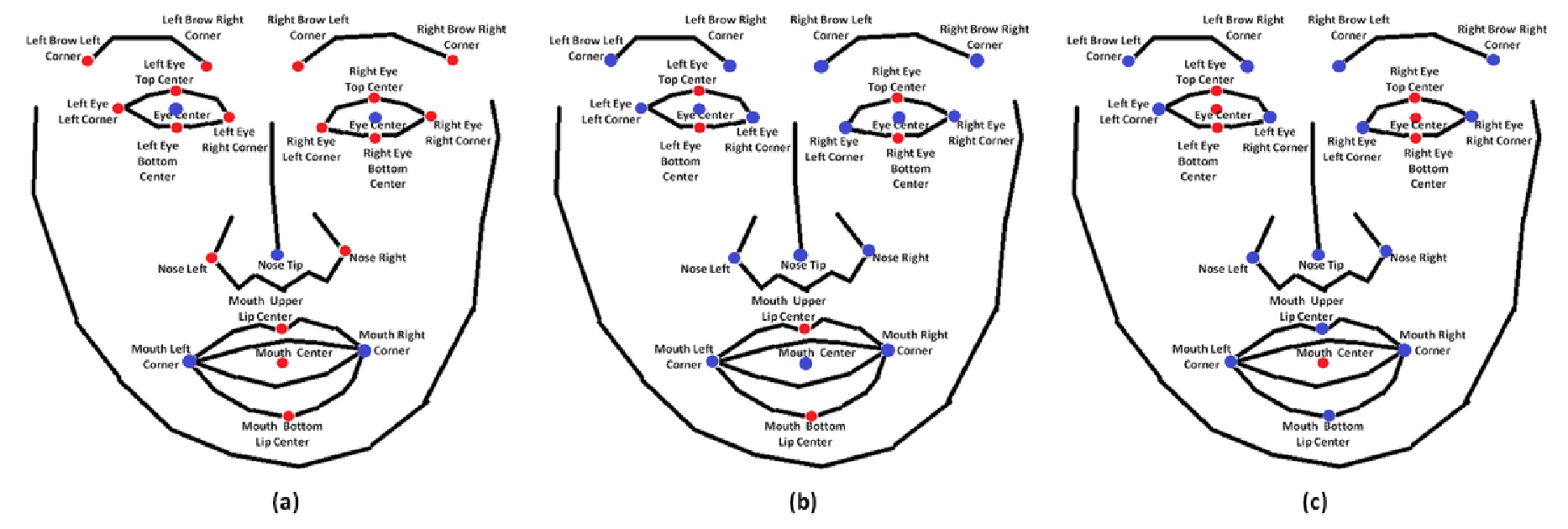}
\caption{Chosen (a)5, (b)16, (c)15 overlapped facial landmarks (blue dots) from 5, 19, 68 points for evaluation.} 
\label{fig:choseface}
\end{figure*}

\subsection{Measure Metric}
\par In our experiment, we adopt four mainly used measure metric: Normalized Mean Error(NME), Cumulative Error Distribution Curve (CED), Area Under the error Curve (AUC) and Failure rate.

\par Normalized Mean Error is calculated using the Euclidean Distance ($L_2$ norm) between estimated points and groundtruth, and being normalized by the distance of two outer eye corners. For each point, landmark-wise normalized error is calculated by:

\begin{equation*}
e_i = \frac{\|x_{(i)}^e - x_{(i)}^g \|_2}{d_{io}} 
\end{equation*}

where, $e_i$ is the i-th error value, $x_{(i)}^e$ is the i-th estimated points, $x_{(i)}^g$ is the i-th ground truth, and $d_{io}$ is IOD, the inter-ocular distance, i.e. Euclidean distance between two outer eye corners.

\par For every face, sample-wise NME is calculated by summarizing the normalized errors of all facial points by: 

\begin{equation*}
e = \sum_{i=1}^n e_i
\end{equation*}

where, $e$ is the error value of face, $n$ is the number of facial landmarks. Since the groudtruth of MOBIO faces contains 22 landmarks, the facial landmark detection methods we choose can detect different numbers of facial feature points (i.e., 5, 19, and 68), in our experiment, we choose the overlapped facial landmarks of groundtruth points and detected points. As shown in Figure \ref{fig:choseface}, there are 5 (in 5), 16 (in 19), and 15 (in 68) overlapped points (blue dots) are used for evaluation.




\par The overall normalized mean error is computed by:
\begin{equation*}
error = \frac{\sum_{i=1}^m e}{m}
\end{equation*}
where, $error$ is the overall normalized mean error, $m$ is the number of faces. 

\par CED is the cumulative distribution function of normalized errors, which evaluates the fraction of facial landmarks changes as error threshold changes. It is a better way to handle outliers. In our experiment, we set the error value threshold as 0.08 and 0.1. We partition the error value range [0, 0.08] or [0, 0.1] into 80 or 100 segments with equal step size 0.001. For each error value point X, the fraction of face images whose error value is <= X is calculated.

\par AUC means the area under the error curve CED:
\begin{equation*}
AUC_\alpha = \int_0^\alpha f(e)de
\end{equation*}
where, $e$ is normalized error, $f(e)$ is cumulative error distribution function, $\alpha$ is the upper bound used to calculate the define integration. In our experiment, $\alpha$ is set as 0.08 and 0.1.

\par Failure Rate is to count the fraction of faces whose error value is greater than error value threshold, in our experiment, 0.08 and 0.1 too.

\section{Experimental Results}
\label{exp}
\par This section, we describe the details of experiment implementation first, including details of each method, and then give a through evaluation result of these method on the generated mobile face images, finally provide a thorough comparison of these methods on other databases, e.g., 300W, AFW, AFLW, COFW.  

\subsection{Implementation Details}
\par Seven facial landmark detection methods are selected. Table \ref{tab:modelinfo} gives the detailed experiment information of these models. Some models \cite{zhang2016joint, feng2017dynamic, he2016facial, zadeh2017convolutional, zhang2014facial} can deal with original images directly, and some others need inputting square faces \cite{wu2017facial, feng2018wing, zhang2018combining} with fixed size. Most models adopt MTCNN as face detector before facial landmark detection. Different models can detect different numbers of faces. So during testing, only the visible landmarks are involved in the evaluation. For each comparison we use the biggest set of overlapping landmarks. For example, Tweaked CNN \cite{wu2017facial} detects 5 landmarks, and the biggest overlapping set with groundtruth (22 landmarks) is 5. WingLoss \cite{feng2018wing} and DAC-CSR \cite{feng2017dynamic} detect 19 landmarks, and finally 16 landmarks are used. PA-CNN \cite{he2016facial}, CE-CLM \cite{zadeh2017convolutional}, ECT \cite{zhang2018combining}, and TCDCN \cite{zhang2014facial} adopt 15 landmarks in 68 for evaluation.


\begin{table*}[!t]
\caption{Information of selected facial landmark models.}
\label{tab:modelinfo}
\begin{tabular}{ccccccc}
\hline\noalign{\smallskip}
\textbf{Method} & \textbf{Face Detector} & \textbf{Input Size} & \textbf{Output Size} & \textbf{\#Detected Landmarks} & \textbf{\#Used Landmarks} & \textbf{\#Detected Faces} \\
\noalign{\smallskip}\hline\noalign{\smallskip}
Tweaked CNN \cite{wu2017facial} & - & 256*256 & 40*40 & 5 & 5 & 20,481 \\
WingLoss \cite{feng2018wing} & MTCNN & 256*256 & 256*256 & 19 & 16 & 20,481  \\
DAC-CSR \cite{feng2017dynamic} & MTCNN & 640*480 & 640*480 & 19 & 16 & 20,481  \\
PA-CNN \cite{he2016facial} & - & 640*480 & 640*480 & 68 & 15 & 19,905 \\
CE-CLM \cite{zadeh2017convolutional} & MTCNN & 640*480 & 640*480 & 68 & 15 & 20,487 \\
ECT \cite{zhang2018combining} & - & 256*256 & 256*256 & 68 & 15 & 20,481 \\
TCDCN \cite{zhang2014facial} & MTCNN & 640*480 & 640*480 & 68 & 15 & 20,481  \\
\noalign{\smallskip}\hline
\end{tabular}
\end{table*}

\subsection{Evaluation on Mobile Still Face Images}
\par
\begin{figure*}[!t]
\center
\includegraphics[width=1.0\linewidth]{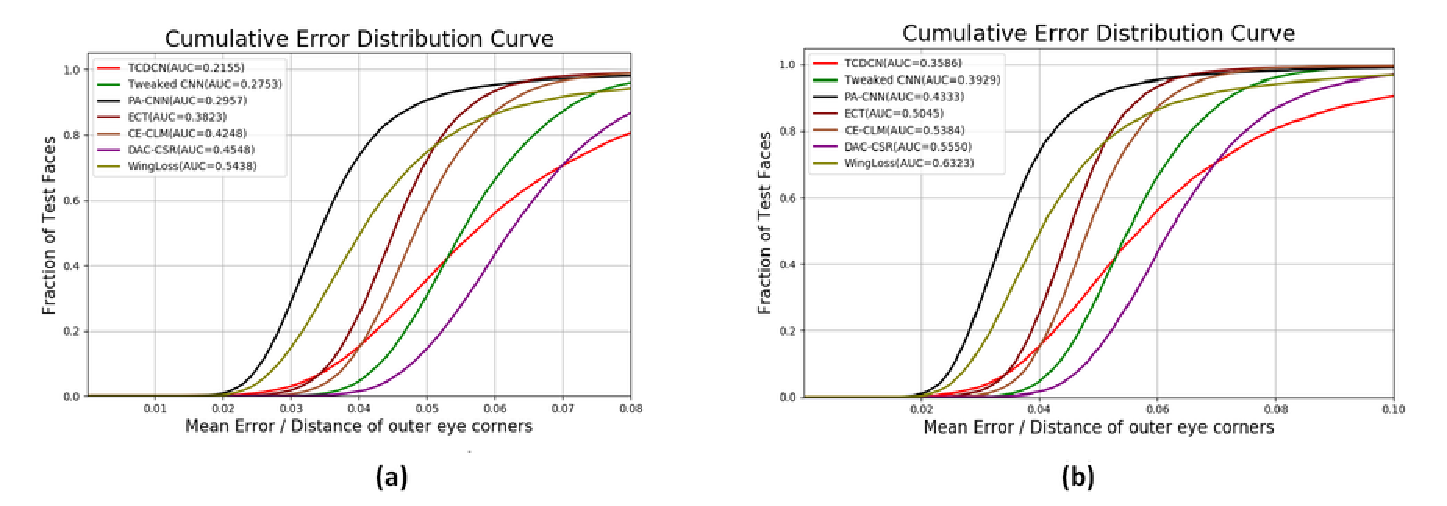}
\caption{CED comparison of all models as error threshold is (a) 0.08, and (b) 0.10.} 
\label{fig:ced}
\end{figure*}

 Seven models are performed facial landmark detection on our generated face data based on MOBIO. Table \ref{tab:perf} provides the normalized mean error, AUC and failure rate when the thresholds are set to 0.08 and 0.1. One can see WingLoss \cite{feng2018wing} gains the lowest mean error and greatest AUC as the threshold is equal to 0.08 and 0.1. TCDCN \cite{zhang2014facial} gains the greatest mean error and the lowest AUC as the threshold is equal to 0.08 and 0.1. CE-CLM \cite{zadeh2017convolutional} obtains the smallest failure rate when the failure rate is defined by the percentage of test images with more than 8\% detection error. ECT \cite{zhang2018combining} obtains the smallest failure rate when the failure rate is defined by the percentage of test images with more than 10\% detection error. Figure \ref{fig:ced} gives the CED curve of all models. Figure \ref{fig:ced}(a) shows the curves with threshold as 0.08, and (b) as 0.1. One can see the performance in Figure \ref{fig:ced}(a) is similar to those in Figure \ref{fig:ced}(b).



\begin{table*}[!t]
\caption{Evaluation results of facial landmark detection on deep models}
\label{tab:perf}
\begin{tabular}{cccccc}
\hline\noalign{\smallskip}
\multirow{2}{*}{Method} & Normalized Mean Error & \multicolumn{2}{c}{Threshold=0.08} & \multicolumn{2}{c}{Threshold=0.10} \\
& ($10^{-2}$) & AUC (\%) & Failure Rate (\%) & AUC (\%) & Failure Rate (\%) \\
\noalign{\smallskip}\hline\noalign{\smallskip}
Tweaked CNN \cite{wu2017facial} & 6.4739049 & 27.533598 & 19.334993 & 39.288243 & 9.462429 \\
WingLoss \cite{feng2018wing} & \textbf{3.8777522} & \textbf{54.384399} & 1.904204 & \textbf{63.232557} & 1.010693 \\
DAC-CSR \cite{feng2017dynamic} & 4.6757547 & 45.475898 & 5.849324 & 55.507959 &	3.251794 \\
PA-CNN \cite{he2016facial} & 5.7171261 & 29.574564 & 4.029736 & 43.333145 & 0.630608 \\
CE-CLM \cite{zadeh2017convolutional} & 4.7493759 & 42.482611 & \textbf{0.990872} & 53.840948 & 0.536926 \\
ECT \cite{zhang2018combining} & 5.0704699 & 38.226405 & 1.079049 & 50.450906 & \textbf{0.502905} \\
TCDCN \cite{zhang2014facial} & 6.5829441 & 21.545304 & 13.290367 & 35.863483 & 3.071139 \\
\noalign{\smallskip}\hline
\end{tabular}
\end{table*}

\par Although there is much ongoing research in computer vision approaches for face alignment, varying evaluation protocols, lack descriptions of critical details and the use of different experimental setting or datasets makes it hard to shed light on how to make an assessment of their cons and pros, and what are the important factors influential to performance. We try our best to make a through performance comparison of the selected methods on our data and other databases, such as 300W, AFLW, AFW, and COFW. 



\par Tweaked CNN \cite{wu2017facial} computes NME values on AFW and AFLW by normalizing the mean distance between predicted to ground truth landmark locations to a percent of the inter-ocular distance. It detects 5 facial feature points for each dataset. 
Tweaked CNN also detects 49 and 68 points on 300W and calculates AUC and failure rate with threshold as 0.1. For 49 points, AUC is 0.817 and failure rate is 1.17\%. For 68 points, the AUC is 0.771 and failure rate is 1.95\%. Both values of AUC are greater than that on our data (39.29\%), and both failure rates are less than that on our data(9.46\%).

\par WingLoss \cite{feng2018wing} is evaluated on AFLW and 300W via calculating NME. For the AFLW dataset, AFLW-Full protocol is adopted, and the width (or height) of the given face bounding box as the normalization term. 1.65\% is gained finally, which is much lower than that on our data. For 300W dataset, the NME uses the inter-pupil distance as the normalization term, and the face images involved in the 300W dataset have been semi-automatically annotated by 68 facial landmarks. The final size of the test set is 689. The test set is further divided into two subsets for evaluation, i.e. the common and challenging subsets. The common subset has 554 face images from the LFPW and HELEN test subsets and the challenging subset constitutes the 135 IBUG face images. 3.27\% is gained finally on Common set which is lower than our faces (3.88\%) 

\par DAC-CSR \cite{feng2017dynamic} is evaluated on AFLW and COFW. For AFLW, 19 landmarks per image without the two ear landmarks are opted. And two protocols (i.e., AFLW-full, AFLW-frontal) are used. AFLW-full uses 4,386 images for test and AFLW-frontal uses 1,165. The performance is measured in terms of the average error, normalized by face size. 2.27\% and 1.81\% are obtained on AFLW-full and AFLW-frontal, which are lower than that on MOBIO (4.68\%).  

\par PA-CNN \cite{he2016facial} evaluates the alignment accuracy on 300w by the mean error, which is measured by the distances between the predicted landmarks and the groundtruth, normalized by the inter-pupil distance. The 300w is divided into three sets, i.e., Common, Challenging and Fullset, and 4.82\%, 9.80\%, and 5.79\% mean errors are gained. In them, the mean error on Common set is lower than ours (5.72\%).

\par CE-CLM \cite{zadeh2017convolutional} is also evaluated on the typical split Common set of 300w by NME, and gains 3.14\% and 2.30\% with outline (68) and without outline (49) separately, which are much lower than ours (4.75\%).

\par ECT \cite{zhang2018combining} evaluated its performance on four databases (300w, AFLW, AFW, and COFW). NME (\%), AUC, and/or Failure Rate (\%) are calculated. The evaluation on 300W consists of two parts. The first part is conducted on the 300W test set provided officially by the 300W competition. The second part of the evaluation is performed on the fullset of 300W which is widely used in the literature. The error is normalized by the distance of outer corners of the eyes. Failure rate is calculated with the threshold set to 0.08 for the normalized point-to-point error. The AUC and failure rate are 45.98\% and 3.17\% with 68 points, which are higher than ours (38.23\%, 1.08\%), and 58.26\% and 1.17\% with 51 points on the test set of 300W competition which are higher too. And the NME are 4.66\%, 7.96\%, and 5.31\% on the Common subset, Challenging subset, and Fullset of 300W. In them, the mean error on Common set is lower than ours (5.07\%).

\par The evaluation on AFLW-PIFA is performed by NME, which are 3.21\% and 3.36\% with 21 and 34 points. Both are lower than ours(5.07\%). 

\par ECT \cite{zhang2018combining} picked out 6 visible landmarks for evaluation on AFW. For NME, the normalized distance is the square root of the bounding box size provided in the AFW dataset. Finally, 2.62\% is obtained, which is lower too (5.07\%).


\par TCDCN \cite{zhang2014facial} is evaluated on 300W, AFLW, AFW and COFW using NME and failure rate. The mean error is measured by the distances between estimated landmarks and the ground truths, normalizing with respect to the inter-ocular distance. Mean error larger than 10\% is reported as a failure. 
And the NME on Common Subset, Challenging Subset, and Fullset of 300W are 4.80\%, 8.60\%, and 5.54\%. In them, the NME on Common Subset and Fullset are lower than ours (6.58\%).   

\par Based on the abovementioned comparison, one can see it is a little bit difficult to tell clearly on which database the selected method perform better due to different measure metrics, and settings. However, in most cases, our mobile face images are more challenging than existing still face images. Our face data can be a new database for facial landmark detection evaluation with 22 facial landmarks as grountruth.

\section{Discussion and Conclusion}
\label{con}
\par MOBIO is a mobile biometrics database captured almost exclusively using mobile phones, which provides a challenging test-bed both for face verification, speaker verification, and bi-modal verification. In this paper, we generate a mobile still face database with 20,600 images based on the MOBIO database and manually label all faces with 22 facial landmarks as groundtruth. Seven state-of-the-art facial landmark detection methods are adopted to evaluate their performance on these 20,600 face images. A thorough analysis about the result and the comparison on other databases are given too. The result shows that our dataset is a pretty challenging one for facial landmark detection.



\section{Acknowledgments}
This work was partly supported by a NSF-CITeR grant and a WV HEPC grant. The authors would like to thank the editors and the anonymous reviewers for the comments and suggestions to improve the manuscript.

\bibliographystyle{IEEEtran}
\bibliography{Facial_landmark_detection_evaluation_on_MOBIO}

\begin{thebibliography}{10}
\providecommand{\url}[1]{#1}
\csname url@samestyle\endcsname
\providecommand{\newblock}{\relax}
\providecommand{\bibinfo}[2]{#2}
\providecommand{\BIBentrySTDinterwordspacing}{\spaceskip=0pt\relax}
\providecommand{\BIBentryALTinterwordstretchfactor}{4}
\providecommand{\BIBentryALTinterwordspacing}{\spaceskip=\fontdimen2\font plus
\BIBentryALTinterwordstretchfactor\fontdimen3\font minus
  \fontdimen4\font\relax}
\providecommand{\BIBforeignlanguage}[2]{{%
\expandafter\ifx\csname l@#1\endcsname\relax
\typeout{** WARNING: IEEEtran.bst: No hyphenation pattern has been}%
\typeout{** loaded for the language `#1'. Using the pattern for}%
\typeout{** the default language instead.}%
\else
\language=\csname l@#1\endcsname
\fi
#2}}
\providecommand{\BIBdecl}{\relax}
\BIBdecl

\bibitem{mccool2012bi}
C.~McCool, S.~Marcel, A.~Hadid, M.~Pietik{\"a}inen, P.~Matejka,
  J.~Cernock{\`y}, N.~Poh, J.~Kittler, A.~Larcher, C.~Levy \emph{et~al.},
  ``Bi-modal person recognition on a mobile phone: using mobile phone data,''
  in \emph{2012 IEEE International Conference on Multimedia and Expo
  Workshops}.\hskip 1em plus 0.5em minus 0.4em\relax IEEE, 2012, pp. 635--640.

\bibitem{liu2017sphereface}
W.~Liu, Y.~Wen, Z.~Yu, M.~Li, B.~Raj, and L.~Song, ``Sphereface: Deep
  hypersphere embedding for face recognition,'' in \emph{Proceedings of the
  IEEE conference on computer vision and pattern recognition}, 2017, pp.
  212--220.

\bibitem{masi2016pose}
I.~Masi, S.~Rawls, G.~Medioni, and P.~Natarajan, ``Pose-aware face recognition
  in the wild,'' in \emph{Proceedings of the IEEE conference on computer vision
  and pattern recognition}, 2016, pp. 4838--4846.

\bibitem{taigman2014deepface}
Y.~Taigman, M.~Yang, M.~Ranzato, and L.~Wolf, ``Deepface: Closing the gap to
  human-level performance in face verification,'' in \emph{Proceedings of the
  IEEE conference on computer vision and pattern recognition}, 2014, pp.
  1701--1708.

\bibitem{yang2017neural}
J.~Yang, P.~Ren, D.~Zhang, D.~Chen, F.~Wen, H.~Li, and G.~Hua, ``Neural
  aggregation network for video face recognition,'' in \emph{Proceedings of the
  IEEE conference on computer vision and pattern recognition}, 2017, pp.
  4362--4371.

\bibitem{fabian2016emotionet}
C.~Fabian Benitez-Quiroz, R.~Srinivasan, and A.~M. Martinez, ``Emotionet: An
  accurate, real-time algorithm for the automatic annotation of a million
  facial expressions in the wild,'' in \emph{Proceedings of the IEEE Conference
  on Computer Vision and Pattern Recognition}, 2016, pp. 5562--5570.

\bibitem{li2017reliable}
S.~Li, W.~Deng, and J.~Du, ``Reliable crowdsourcing and deep
  locality-preserving learning for expression recognition in the wild,'' in
  \emph{Proceedings of the IEEE conference on computer vision and pattern
  recognition}, 2017, pp. 2852--2861.

\bibitem{walecki2016copula}
R.~Walecki, O.~Rudovic, V.~Pavlovic, and M.~Pantic, ``Copula ordinal regression
  for joint estimation of facial action unit intensity,'' in \emph{Proceedings
  of the IEEE Conference on Computer Vision and Pattern Recognition}, 2016, pp.
  4902--4910.

\bibitem{zeng2008survey}
Z.~Zeng, M.~Pantic, G.~I. Roisman, and T.~S. Huang, ``A survey of affect
  recognition methods: Audio, visual, and spontaneous expressions,'' \emph{IEEE
  transactions on pattern analysis and machine intelligence}, vol.~31, no.~1,
  pp. 39--58, 2008.

\bibitem{dou2017end}
P.~Dou, S.~K. Shah, and I.~A. Kakadiaris, ``End-to-end 3d face reconstruction
  with deep neural networks,'' in \emph{Proceedings of the IEEE Conference on
  Computer Vision and Pattern Recognition}, 2017, pp. 5908--5917.

\bibitem{kittler20163d}
J.~Kittler, P.~Huber, Z.-H. Feng, G.~Hu, and W.~Christmas, ``3d morphable face
  models and their applications,'' in \emph{International Conference on
  Articulated Motion and Deformable Objects}.\hskip 1em plus 0.5em minus
  0.4em\relax Springer, 2016, pp. 185--206.

\bibitem{huber2016real}
P.~Huber, P.~Kopp, W.~Christmas, M.~R{\"a}tsch, and J.~Kittler, ``Real-time 3d
  face fitting and texture fusion on in-the-wild videos,'' \emph{IEEE Signal
  Processing Letters}, vol.~24, no.~4, pp. 437--441, 2016.

\bibitem{hu2017efficient}
G.~Hu, F.~Yan, J.~Kittler, W.~Christmas, C.~H. Chan, Z.~Feng, and P.~Huber,
  ``Efficient 3d morphable face model fitting,'' \emph{Pattern Recognition},
  vol.~67, pp. 366--379, 2017.

\bibitem{roth2016adaptive}
J.~Roth, Y.~Tong, and X.~Liu, ``Adaptive 3d face reconstruction from
  unconstrained photo collections,'' in \emph{Proceedings of the IEEE
  Conference on Computer Vision and Pattern Recognition}, 2016, pp. 4197--4206.

\bibitem{koppen2018gaussian}
P.~Koppen, Z.-H. Feng, J.~Kittler, M.~Awais, W.~Christmas, X.-J. Wu, and H.-F.
  Yin, ``Gaussian mixture 3d morphable face model,'' \emph{Pattern
  Recognition}, vol.~74, pp. 617--628, 2018.

\bibitem{demirkus2015hierarchical}
M.~Demirkus, D.~Precup, J.~J. Clark, and T.~Arbel, ``Hierarchical temporal
  graphical model for head pose estimation and subsequent attribute
  classification in real-world videos,'' \emph{Computer Vision and Image
  Understanding}, vol. 136, pp. 128--145, 2015.

\bibitem{zhu2012face}
X.~Zhu and D.~Ramanan, ``Face detection, pose estimation, and landmark
  localization in the wild,'' in \emph{2012 IEEE conference on computer vision
  and pattern recognition}.\hskip 1em plus 0.5em minus 0.4em\relax IEEE, 2012,
  pp. 2879--2886.

\bibitem{ding2013facial}
X.~Ding, W.-S. Chu, F.~De~la Torre, J.~F. Cohn, and Q.~Wang, ``Facial action
  unit event detection by cascade of tasks,'' in \emph{Proceedings of the IEEE
  international conference on computer vision}, 2013, pp. 2400--2407.

\bibitem{martinez2016advances}
B.~Martinez and M.~F. Valstar, ``Advances, challenges, and opportunities in
  automatic facial expression recognition,'' in \emph{Advances in face
  detection and facial image analysis}.\hskip 1em plus 0.5em minus 0.4em\relax
  Springer, 2016, pp. 63--100.

\bibitem{sariyanidi2014automatic}
E.~Sariyanidi, H.~Gunes, and A.~Cavallaro, ``Automatic analysis of facial
  affect: A survey of registration, representation, and recognition,''
  \emph{IEEE transactions on pattern analysis and machine intelligence},
  vol.~37, no.~6, pp. 1113--1133, 2014.

\bibitem{liu2015deep}
Z.~Liu, P.~Luo, X.~Wang, and X.~Tang, ``Deep learning face attributes in the
  wild,'' in \emph{Proceedings of the IEEE international conference on computer
  vision}, 2015, pp. 3730--3738.

\bibitem{cristinacce2006feature}
D.~Cristinacce and T.~F. Cootes, ``Feature detection and tracking with
  constrained local models.'' in \emph{Bmvc}, vol.~1, no.~2.\hskip 1em plus
  0.5em minus 0.4em\relax Citeseer, 2006, p.~3.

\bibitem{cao2014face}
X.~Cao, Y.~Wei, F.~Wen, and J.~Sun, ``Face alignment by explicit shape
  regression,'' \emph{International Journal of Computer Vision}, vol. 107,
  no.~2, pp. 177--190, 2014.

\bibitem{xiong2013supervised}
X.~Xiong and F.~De~la Torre, ``Supervised descent method and its applications
  to face alignment,'' in \emph{Proceedings of the IEEE conference on computer
  vision and pattern recognition}, 2013, pp. 532--539.

\bibitem{hassner2015effective}
T.~Hassner, S.~Harel, E.~Paz, and R.~Enbar, ``Effective face frontalization in
  unconstrained images,'' in \emph{Proceedings of the IEEE Conference on
  Computer Vision and Pattern Recognition}, 2015, pp. 4295--4304.

\bibitem{zadeh2016mosi}
A.~Zadeh, R.~Zellers, E.~Pincus, and L.-P. Morency, ``Mosi: multimodal corpus
  of sentiment intensity and subjectivity analysis in online opinion videos,''
  \emph{arXiv preprint arXiv:1606.06259}, 2016.

\bibitem{sun2014deep}
Y.~Sun, X.~Wang, and X.~Tang, ``Deep learning face representation from
  predicting 10,000 classes,'' in \emph{Proceedings of the IEEE conference on
  computer vision and pattern recognition}, 2014, pp. 1891--1898.

\bibitem{ren2014face}
S.~Ren, X.~Cao, Y.~Wei, and J.~Sun, ``Face alignment at 3000 fps via regressing
  local binary features,'' in \emph{Proceedings of the IEEE Conference on
  Computer Vision and Pattern Recognition}, 2014, pp. 1685--1692.

\bibitem{zhang2014coarse}
J.~Zhang, S.~Shan, M.~Kan, and X.~Chen, ``Coarse-to-fine auto-encoder networks
  (cfan) for real-time face alignment,'' in \emph{European conference on
  computer vision}.\hskip 1em plus 0.5em minus 0.4em\relax Springer, 2014, pp.
  1--16.

\bibitem{zhu2015face}
S.~Zhu, C.~Li, C.~Change~Loy, and X.~Tang, ``Face alignment by coarse-to-fine
  shape searching,'' in \emph{Proceedings of the IEEE conference on computer
  vision and pattern recognition}, 2015, pp. 4998--5006.

\bibitem{zhang2015learning}
Z.~Zhang, P.~Luo, C.~C. Loy, and X.~Tang, ``Learning deep representation for
  face alignment with auxiliary attributes,'' \emph{IEEE transactions on
  pattern analysis and machine intelligence}, vol.~38, no.~5, pp. 918--930,
  2015.

\bibitem{cootes1995active}
T.~F. Cootes, C.~J. Taylor, D.~H. Cooper, and J.~Graham, ``Active shape
  models-their training and application,'' \emph{Computer vision and image
  understanding}, vol.~61, no.~1, pp. 38--59, 1995.

\bibitem{cootes2001active}
T.~F. Cootes, G.~J. Edwards, and C.~J. Taylor, ``Active appearance models,''
  \emph{IEEE Transactions on Pattern Analysis \& Machine Intelligence}, no.~6,
  pp. 681--685, 2001.

\bibitem{wu2017simultaneous}
Y.~Wu, C.~Gou, and Q.~Ji, ``Simultaneous facial landmark detection, pose and
  deformation estimation under facial occlusion,'' in \emph{Proceedings of the
  IEEE conference on computer vision and pattern recognition}, 2017, pp.
  3471--3480.

\bibitem{feng2017face}
Z.-H. Feng, J.~Kittler, M.~Awais, P.~Huber, and X.-J. Wu, ``Face detection,
  bounding box aggregation and pose estimation for robust facial landmark
  localisation in the wild,'' in \emph{Proceedings of the IEEE conference on
  computer vision and pattern recognition workshops}, 2017, pp. 160--169.

\bibitem{wu2016constrained}
Y.~Wu and Q.~Ji, ``Constrained joint cascade regression framework for
  simultaneous facial action unit recognition and facial landmark detection,''
  in \emph{Proceedings of the IEEE conference on computer vision and pattern
  recognition}, 2016, pp. 3400--3408.

\bibitem{feng2014random}
Z.-H. Feng, P.~Huber, J.~Kittler, W.~Christmas, and X.-J. Wu, ``Random
  cascaded-regression copse for robust facial landmark detection,'' \emph{IEEE
  Signal Processing Letters}, vol.~22, no.~1, pp. 76--80, 2014.

\bibitem{sun2013deep}
Y.~Sun, X.~Wang, and X.~Tang, ``Deep convolutional network cascade for facial
  point detection,'' in \emph{Proceedings of the IEEE conference on computer
  vision and pattern recognition}, 2013, pp. 3476--3483.

\bibitem{feng2015cascaded}
Z.-H. Feng, G.~Hu, J.~Kittler, W.~Christmas, and X.-J. Wu, ``Cascaded
  collaborative regression for robust facial landmark detection trained using a
  mixture of synthetic and real images with dynamic weighting,'' \emph{IEEE
  Transactions on Image Processing}, vol.~24, no.~11, pp. 3425--3440, 2015.

\bibitem{feng2018wing}
Z.-H. Feng, J.~Kittler, M.~Awais, P.~Huber, and X.-J. Wu, ``Wing loss for
  robust facial landmark localisation with convolutional neural networks,'' in
  \emph{Proceedings of the IEEE Conference on Computer Vision and Pattern
  Recognition}, 2018, pp. 2235--2245.

\bibitem{jourabloo2015pose}
A.~Jourabloo and X.~Liu, ``Pose-invariant 3d face alignment,'' in
  \emph{Proceedings of the IEEE International Conference on Computer Vision},
  2015, pp. 3694--3702.

\bibitem{lee2015face}
D.~Lee, H.~Park, and C.~D. Yoo, ``Face alignment using cascade gaussian process
  regression trees,'' in \emph{Proceedings of the IEEE Conference on Computer
  Vision and Pattern Recognition}, 2015, pp. 4204--4212.

\bibitem{zhang2014facial}
Z.~Zhang, P.~Luo, C.~C. Loy, and X.~Tang, ``Facial landmark detection by deep
  multi-task learning,'' in \emph{European conference on computer
  vision}.\hskip 1em plus 0.5em minus 0.4em\relax Springer, 2014, pp. 94--108.

\bibitem{sagonas2013300}
C.~Sagonas, G.~Tzimiropoulos, S.~Zafeiriou, and M.~Pantic, ``300 faces
  in-the-wild challenge: The first facial landmark localization challenge,'' in
  \emph{Proceedings of the IEEE International Conference on Computer Vision
  Workshops}, 2013, pp. 397--403.

\bibitem{koestinger2011annotated}
M.~Koestinger, P.~Wohlhart, P.~M. Roth, and H.~Bischof, ``Annotated facial
  landmarks in the wild: A large-scale, real-world database for facial landmark
  localization,'' in \emph{2011 IEEE international conference on computer
  vision workshops (ICCV workshops)}.\hskip 1em plus 0.5em minus 0.4em\relax
  IEEE, 2011, pp. 2144--2151.

\bibitem{belhumeur2013localizing}
P.~N. Belhumeur, D.~W. Jacobs, D.~J. Kriegman, and N.~Kumar, ``Localizing parts
  of faces using a consensus of exemplars,'' \emph{IEEE transactions on pattern
  analysis and machine intelligence}, vol.~35, no.~12, pp. 2930--2940, 2013.

\bibitem{le2012interactive}
V.~Le, J.~Brandt, Z.~Lin, L.~Bourdev, and T.~S. Huang, ``Interactive facial
  feature localization,'' in \emph{European conference on computer
  vision}.\hskip 1em plus 0.5em minus 0.4em\relax Springer, 2012, pp. 679--692.

\bibitem{burgos2013robust}
X.~P. Burgos-Artizzu, P.~Perona, and P.~Doll{\'a}r, ``Robust face landmark
  estimation under occlusion,'' in \emph{Proceedings of the IEEE International
  Conference on Computer Vision}, 2013, pp. 1513--1520.

\bibitem{messer1999xm2vtsdb}
K.~Messer, J.~Matas, J.~Kittler, J.~Luettin, and G.~Maitre, ``Xm2vtsdb: The
  extended m2vts database,'' in \emph{Second international conference on audio
  and video-based biometric person authentication}, vol. 964, 1999, pp.
  965--966.

\bibitem{marcel2010results}
S.~Marcel, C.~McCool, P.~Mat{\v{e}}jka, T.~Ahonen, J.~{\v{C}}ernock{\`y},
  S.~Chakraborty, V.~Balasubramanian, S.~Panchanathan, C.~H. Chan, J.~Kittler
  \emph{et~al.}, ``On the results of the first mobile biometry (mobio) face and
  speaker verification evaluation,'' in \emph{International Conference on
  Pattern Recognition}.\hskip 1em plus 0.5em minus 0.4em\relax Springer, 2010,
  pp. 210--225.

\bibitem{feng2017dynamic}
Z.-H. Feng, J.~Kittler, W.~Christmas, P.~Huber, and X.-J. Wu, ``Dynamic
  attention-controlled cascaded shape regression exploiting training data
  augmentation and fuzzy-set sample weighting,'' in \emph{Proceedings of the
  IEEE Conference on Computer Vision and Pattern Recognition}, 2017, pp.
  2481--2490.

\bibitem{he2016facial}
K.~He and X.~Xue, ``Facial landmark localization by part-aware deep
  convolutional network,'' in \emph{Pacific Rim Conference on
  Multimedia}.\hskip 1em plus 0.5em minus 0.4em\relax Springer, 2016, pp.
  22--31.

\bibitem{zadeh2017convolutional}
A.~Zadeh, T.~Baltrusaitis, and L.-P. Morency, ``Convolutional experts network
  for facial landmark detection,'' in \emph{Proceedings of the International
  Conference on Computer Vision \& Pattern Recognition (CVPRW),
  Faces-in-the-wild Workshop/Challenge}, vol.~3, no.~5, 2017, p.~6.

\bibitem{wu2017facial}
Y.~Wu, T.~Hassner, K.~Kim, G.~Medioni, and P.~Natarajan, ``Facial landmark
  detection with tweaked convolutional neural networks,'' \emph{IEEE
  transactions on pattern analysis and machine intelligence}, vol.~40, no.~12,
  pp. 3067--3074, 2017.

\bibitem{zhang2018combining}
H.~Zhang, Q.~Li, Z.~Sun, and Y.~Liu, ``Combining data-driven and model-driven
  methods for robust facial landmark detection,'' \emph{IEEE Transactions on
  Information Forensics and Security}, vol.~13, no.~10, pp. 2409--2422, 2018.

\bibitem{zhang2016joint}
K.~Zhang, Z.~Zhang, Z.~Li, and Y.~Qiao, ``Joint face detection and alignment
  using multitask cascaded convolutional networks,'' \emph{IEEE Signal
  Processing Letters}, vol.~23, no.~10, pp. 1499--1503, 2016.

\end{thebibliography}

\end{document}